\documentclass[10pt, a4paper]{article}
\usepackage{lrec}
\usepackage{graphicx}
\usepackage{tabularx}
\usepackage{soul}

\usepackage{epstopdf}
\usepackage[utf8]{inputenc}
\usepackage{amssymb}
\usepackage{booktabs}
\usepackage{xstring}
\usepackage{titlesec}
\titleformat{\section}{\normalfont\large\bf\center}{\thesection.}{1em}{}
\titleformat{\subsection}{\normalfont\SmallTitleFont\bf\raggedright}{\thesubsection.}{1em}{}
\titleformat{\subsubsection}{\normalfont\normalsize\bf\raggedright}{\thesubsubsection.}{1em}{}
\renewcommand\thesection{\arabic{section}}
\renewcommand\thesubsection{\thesection.\arabic{subsection}}
\renewcommand\thesubsubsection{\thesubsection.\arabic{subsubsection}}

\title{Project PIAF:\\Building a Native French Question-Answering Dataset}

\name{Rachel Keraron$^{\star\ast}$\thanks{$\ast$: equal contribution}, 
Guillaume Lancrenon$^{\ddagger\ast}$, Mathilde Bras$^{\ddagger\ast}$,\\
        {\bf \large Frédéric Allary$^{\star}$, Gilles Moyse$^{\star}$, 
        Thomas Scialom$^{\star \diamond}$}\\
        {\bf \large Edmundo-Pavel Soriano-Morales$^{\ddagger}$, Jacopo Staiano$^{\star}$}}

\address{${\star}$ reciTAL, Paris (France)\\ ${\ddagger}$Etalab, DINUM, Prime Minister's Office, Paris (France) \\
         ${\diamond}$ Sorbonne Universit\'e, CNRS, LIP6, F-75005 Paris, France\\
         \{rachel, frederic, gilles, thomas, jacopo\}@recital.ai\\
         \{guillaume.lancrenon, mathilde.bras, pavel.soriano\}@data.gouv.fr\\}

\usepackage[hyperfootnotes=false,colorlinks=true,citecolor=black,urlcolor=blue]{hyperref}

\abstract{
Motivated by the lack of data for non-English languages, in particular for the evaluation of downstream tasks such as Question Answering, we present a participatory effort to collect a native French Question Answering Dataset. Furthermore, we describe and publicly release the annotation tool developed for our collection effort, along with the data obtained and preliminary baselines.\\ \newline  }

\begin{document}

\maketitleabstract

\section{Introduction}

Along with the availability of massive amounts of data, the increase in computational power has in recent years allowed the development of Deep Learning techniques, leading to significant advancements in the fields of Computer Vision (CV), and Natural Language Processing (NLP), among others. Visual information can, to some extent, be considered to generalize across cultures in many real world applications; in contrast, having to deal with languages, NLP applications are naturally bound to language specificities.

Over the years, the NLP community has produced several resources to tackle tasks we call, for simplicity, \emph{upstream} (such as Part-Of-Speech tagging, Dependency Parsing, etc.), targeting multiple languages and enabling the construction of effective automated systems. Still, for tasks we refer to as \emph{downstream}, i.e. those which enable the development of value-added end products such as Question Answering (QA) or Conversational Agents, the current state-of-the-art approaches require massive amounts of annotated data which are almost exclusively available in English. 
Notable exceptions include Machine Translation, for which abundant parallel corpora have been built from resources such as the European parliamentary proceedings, or Language Modeling -- which can be tackled in a self-supervised manner, hence only requiring massive amounts of text in the target language(s).
Such asymmetry has recently been acknowledged by a resolution of the EU Parliament.\footnote{European Parliament resolution of 11 September 2018 on language equality in the digital age. \url{http://www.europarl.europa.eu/sides/getDoc.do?pubRef=-//EP//TEXT+TA+P8-TA-2018-0332+0+DOC+XML+V0//EN}}

In this work, we focus on collecting Question Answering data in French in a participatory setup: we describe the data collection protocol adopted, report descriptive statistics and baselines, and provide details on the implementation of the open source annotation tool we developed. Such tool allows volunteers to participate in crowdsourced QA dataset collection campaigns.

In summary, we make the following contributions: 
\begin{enumerate}
    \item we develop and release a novel annotation tool to collect large-scale QA data in a participatory scenario;
    \item we release a native French QA dataset;
    \item we provide baselines using state-of-the-art methodologies.
\end{enumerate}

\section{Related Work}
\label{sec:related_work}

Several datasets for QA have been recently produced. The Stanford Question Answering Dataset (SQuAD) \cite{rajpurkar2016squad} consists of 100,000+ questions posed by crowdworkers on a set of Wikipedia articles. The answer to each question is a segment of text from the corresponding article. Its latest version includes circa 50k additional unanswerable questions \cite{rajpurkar2018know}.

Targeting open-domain QA, WikiQA \cite{yang2015wikiqa} consists of Bing queries and corresponding answer sentences taken from Wikipedia articles. CoQA \cite{reddy2018coqa} focuses on conversational aspects, compounding to 127,000+ questions with answers collected from 8000+ conversations. Each conversation is collected by pairing two crowdworkers to chat about a passage in the form of questions and answers. Based on CNN news articles, NewsQA \cite{trischler2017newsqa} consists of 120k Q\&A pairs, including unanswerable questions, collected using a 3-stage, siloed process; crowdworkers are split into 3 groups: Questioners, who see only an article's headline and highlights to produce a question; Answerers, who see the question and the full article, then select an answer passage; and Validators, who see the article, the question, and a set of answers that they rank. 

Recognizing that several pieces of information often jointly imply another fact, the two datasets provided by QAngaroo \cite{welbl2018constructing} target Multi-Hop (or Multi-Step) QA, i.e. the goal is to answer text understanding queries by combining multiple facts that are spread across different documents. The provided datasets, WikiHop and MedHop target, respectively, open-domain and domain-specific QA, with the latter focusing on interactions between pairs of drugs in the medical domain. Also targeting Multi-Hop reasoning, HotpotQA \cite{yang2018hotpotqa} includes 113k Wikipedia-based question-answer pairs where the questions require finding and reasoning over multiple supporting documents to answer.

TriviaQA \cite{joshi2017triviaqa} includes over 650k question-answer-evidence triples, with questions originating from trivia enthusiasts independent of the evidence documents, automatically gathered and hence noisy. In addition, it provides a clean, human-annotated subset of 1975 question-document-answer triples whose documents are certified to contain all facts required to answer the questions.

The Yahoo! Answers datasets, available for English and French under restrictive non-commercial licenses\footnote{\url{https://webscope.sandbox.yahoo.com/catalog.php?datatype=l}}, include question-answer pairs obtained through the Yahoo! Answers service. No supporting passage for an answer is provided, although links to relevant web pages might be included in the more elaborate answers. Available also under a non-commercial license, InsuranceQA \cite{feng2015applying} is a domain-specific QA dataset targeting the insurance domain.

\subsection*{Do We Need Native Non-English Data?} 
In short, \emph{yes}. 

From the above mentioned data collection efforts, we see that with the exception of Yahoo! Answers, which does not provide supporting evidence for the answers, no large-scale  QA dataset is available for languages other than English. Current approaches based on transfer learning and multilingual model pretraining allow to build models able to deal with non-English. In other words, one can effectively fine-tune a pretrained multilingual language model (e.g. \cite{devlin2019bert}) on English QA data. Nonetheless, as observed by \cite{lewis2019mlqa}, native \emph{evaluation} data for the targeted language is a \emph{must-have} in order to measure progress on the task for a given language. \cite{lewis2019mlqa} thus released to the community an aligned multilingual evaluation corpus containing QA data in 7 languages (English, Arabic, German, Spanish, Hindi, Vietnamese and Simplified Chinese).

Recent research~\cite{xcmrc2019} on Chinese language (arguably, the other high-resource language\footnote{In terms of publicly available text and corpora.} along with English) indicates that, while translation-based method and multilingual approaches can obtain reasonable performances, there exists a large margin for improvements. To allow the research community to tackle those issues, it is thus desirable to obtain comparable data for the task, natively for the language of choice. Efforts in this direction have focused on Chinese~\cite{cui2018span}, Korean~\cite{lee-etal-2018-semi,2018korquad}. 

Hence, project PIAF\footnote{\url{https://piaf.etalab.studio/}} (\textit{For a French-language AI} or \textit{Pour une IA Francophone} in French) focuses on filling this gap for French, starting from the Question-Answering use case. In a first stage of the collection effort, we prioritize quality over quantity: we conducted several in-place \emph{annotathons} wherein volunteers are accompanied by the PIAF team. As showing up to the meetings was usually motivated by curiosity towards AI topics, the presence of the PIAF team served to increase engagement and allowed to provide participants with basic knowledge on how AI models can be built. We provide more details on \emph{annotathons} in Section~\ref{sec:crowdsourcing}. Concurrently with our public effort, another team was collecting native French QA data, unbeknownst to us: we provide a comparison with FQUAD \cite{d2020fquad} in Section~\ref{sec:analysis}.

\section{Protocol}
\label{sec:Protocol}

Consistently with the recent works targeting QA data collections in non-English languages, we focus on \emph{extractive} Question Answering and adapt the protocol used for SQuAD: we collect samples in the form of triplets \emph{$\{P, Q, A\}$}, in which the answer $A$ to a question $Q$ is contained in the paragraph $P$.\footnote{The presence of \emph{unanswerable} question marks the difference between SQuADv1.1. and SQuADv2.0.}

The great majority ($\sim 95\%$) of the Wikipedia articles used to build the SQuAD dataset have a version on the French Wikipedia, making an article-by-article replication of the collection process feasible. Still, such approach suffers from several downsides: for instance, cultural differences can cause significant divergence between the English and French versions of the same Wikipedia article, both in terms of data availability and reliability. Taking the Wikipedia article on the 2016 edition of the Super Bowl (present in the SQuAD dataset) as an example, one can easily notice how the contents significantly differ between its \href{https://en.wikipedia.org/wiki/Super_Bowl_50}{English} and \href{https://en.wikipedia.org/wiki/Super_Bowl_50}{French} versions.

Thus, we select relevant articles the French version of Wikipedia, using the same relevance metric as used in SQuAD (\emph{i.e.} PageRank); we segment those articles in smaller paragraphs, and collect sets of question/answer pairs corresponding to those paragraphs.

Two collection settings are envisioned: the first, \emph{certified}, restricted to volunteers participating in live \emph{annotathons} (see Section~\ref{sec:crowdsourcing}); the second, \emph{open}, wherein we will keep instances of the annotation platform open to the public web and engage the community to contribute to the collection, using a similar approach as that of Mozilla Common Voice.\footnote{\url{https://voice.mozilla.org}}

The collected data is publicly released under CC-BY-SA license.\footnote{PIAF data is also available through the HuggingFace \href{https://github.com/huggingface/nlp}{\texttt{nlp}} library.}

\subsection*{Source Articles Selection}
For SQuAD, the authors used the highest-ranking, in terms of PageRank, 10k articles from the English Wikipedia. Nonetheless, when applying the same threshold on the French version, we noticed significant differences in terms of structural properties between French and English Wikipedia. For instance, we observed the massive presence of pages referring to \emph{years} on the French version, a characteristic that the top-10k subsample of the English Wikipedia does not exhibit. After manual inspection, the editing practices seem to differ between the French and English communities: for instance, while the latter do not link all \emph{year} mentions to the actual page dedicated to that year, the French Wikipedia editors seem to systematically do so --- a fact that boosts the PageRank score of such articles and therefore explains their abundant presence in the French top-10k subsample.

To account for such practices, an extensive manual inspection step on random samples from the French Wikipedia allowed us to identify the following filtering criteria. 

At section level, we discard those with the following titles: \texttt{Voir aussi} \emph{(See also)}, \texttt{Articles connexe} \emph{(Related Articles)}, \texttt{Liens externes} \emph{(External links)}, \texttt{Notes et références} \emph{(References)}. Those sections most often contain only bullet lists with structured rather than textual information.

At article level, we discard those containing a section titled \texttt{Événements} \emph{(Events)}: those articles are about years, as in the aforementioned example, their structure is not apt to extractive QA since they mostly consist of lists of events (a date, followed by a very short text). Further, for concerns on data quality we discard articles falling in two Wikipedia categories: \texttt{Catégorie:Wikipédia:ébauche} \emph{(Draft)} and  \texttt{Catégorie:Homonymie} \emph{(Disambiguation)}.

To summarize, we operate as follows:
\begin{itemize}
    \item gather the top-25k articles in terms of PageRank;\footnote{We adapted the code from Project Nayuki \url{https://www.nayuki.io/page/computing-wikipedias-internal-pageranks}}
    \item discard the articles matching the above criteria;
    \item set a min-max char limit on the paragraph length ($min=500; max=1000$);
    \item filter out articles with less than 5 paragraphs.
\end{itemize}

Compared to the English SQuAD, we hence obtain annotated QA data on more articles, with less paragraphs per article, and a comparable length.

\subsubsection*{Category Distribution}

\begin{table*}
\begin{center}
\begin{tabular}{lcccccccr}
      &\textbf{Arts}&\textbf{Geography}&\textbf{History}&\textbf{Religion}&\textbf{Sciences}&\textbf{Society/Misc.}&\textbf{Sport}&\emph{Total}\\
      \hline
      Certified & 52 & 52& 57& 16 & 52 & 36 & 26&\emph{291}\\
      \emph{\hphantom{.1pt}PIAFv1.0} & \emph{30} & \emph{32} & \emph{38} & \emph{12} & \emph{36} & \emph{25} & \emph{18} &\emph{191}\\
      Open & 228 & 155& 233& 85 & 190 & 167 & 97&\emph{1155}\\
      \hline
      Total & 280 & 207& 290& 101 & 242 & 203 & 123&\emph{1446}\\
      \hline

\end{tabular}
\caption{Category distribution of source articles in \emph{Certified} (a subset of which is released as \emph{PIAFv1.0}) and \emph{Open} splits.}
\label{tab:categories}
\end{center}
\end{table*}
We associated each article to the main category it belongs to. To do so, we first analysed the category/project trees associated to each Wikipedia article. We deemed such information as not directly exploitable, given the presence of several ambiguities and inconsistencies. Nonetheless, such analysis allowed us to shortlist a subset of categories which were most represented. We then proceeded to a manual annotation step to associate each article to the most relevant category. The category information would be instrumental as a factor for engagement, in our participatory scenario: the volunteers have control over the general category they want to contribute to. In Table~\ref{tab:categories} we show the distribution of articles per category.

\subsection*{Continuous Evaluation}

It is of utmost importance to drive the users to produce challenging questions. In the SQuAD collection interface, the user was reminded to avoid using the same words/phrases as the paragraph while writing a question, via a text message displayed on screen. In our collection interface, we provide annotators with examples of good and bad questions. 

In order to check the quality of the collected data, we execute both manual and automatic evaluations throughout the data collection process, on a rolling basis (\emph{i.e.} as the collected dataset grows), as in the original SQuAD paper. 

For manual evaluation, we analyzed 191 questions from the collected \emph{certified} data (\emph{i.e.} one randomly sampled triplet per article), and assign scores according to the dimensions defined in the SQuAD paper -- see Table~\ref{tab:manual}.
Conversely, we compute automatically, on the ensemble of the data and on a rolling basis, scores for 
\emph{syntactic divergence} and \emph{lexical variation}. 
Manual and automatic evaluation results for \emph{PIAFv1.0} are discussed in Section~\ref{sec:analysis}.

Further, we incrementally measure the performances obtained by state-of-the-art multilingual QA systems, under several experimental setups.

\begin{table*}
\begin{center}
\begin{tabular}{lcccc}
      \hline
      \textbf{Features}&\textbf{Daemo}&\textbf{cdQA}&\textbf{QA-Turk}&\textbf{PIAFAnno}\\
      \hline
      Multiple users & \checkmark &&& \checkmark\\
      Modern interface &&&& \checkmark\\
      Open source & \checkmark &\checkmark&\checkmark& \checkmark\\
      SQuAD compatible & \checkmark &\checkmark&\checkmark&\checkmark\\
      Actively developed &&\checkmark&\checkmark& \checkmark\\
      \hline

\end{tabular}
\caption{QA annotation tools and their supported features.}
\label{tab:annotation_tools}
\end{center}
\end{table*}

\section{Annotation Platform}
Effective annotation frameworks are essential  for  building  language- and/or domain-specific NLP corpora. 
While multiple QA datasets have recently been produced(see Section \ref{sec:related_work}), there is still a lack of annotation frameworks which are open and accessible to the community.
The main reasons lie in the usage of proprietary software, the deep link with crowd-sourcing marketplaces (e.g. Mechanical Turk), and to the lack of minimal features enabling QA data collection. 
We thus present PIAFAnno, a browser-based, mobile-friendly, QA annotation tool. 
PIAFAnno was created to meet the following constraints:
\begin{enumerate}
\item \textit{Web-based crowd-sourcing platform}: allowing for a distributed, large-scale contribution.
\item \textit{User and contribution management}: supporting different roles among the annotators, as well as keeping control of the progress and the quality of the contributions.
\item \textit{Modern interface}: our collection protocol is centered on voluntary participation. Therefore, the workflow should be pleasing and engaging for the annotators.
\item \textit{SQuAD compatible}: to make the data quickly actionable by the community, the input and output formats follow the SQuAD format. The SQuAD annotation flow is respected (\emph{i.e.} creating a number of questions for each paragraph of a single document).
\item \textit{Permissive and open-source license}: making our tool reusable by the community as a part of our commitment to open source and data.
\end{enumerate}

\subsection{Existing Platforms}
Before launching our development effort, we reviewed existing open-source platforms that could fit for our scenario.
Unfortunately, while there are several sequence-annotation tools, such as \textit{brat} \cite{stenertop2012brat}, \textit{WebAnno} \cite{muhie2013webanno}, or \textit{Doccano}\footnote{We actually began working on implementing a QA module for Doccano but due to a lack of time for pull-requests reviews and other technical details we finally decided to roll out our own solution from scratch.} \cite{nakayama2018doccano},  we identified, to the best of our efforts, only three candidates as open-source, web-based, QA annotation platforms (shown in Table \ref{tab:annotation_tools}). We aim to fill this gap with PIAFAnno.

The crowd-sourcing platform used in the original SQuAD paper \cite{rajpurkar2016squad}, \textit{Daemo} \cite{gaikwad2015daemo}, is publicly available and open-source.\footnote{\url{https://github.com/crowdresearch/daemo}} Nonetheless, its development seems to have stalled and the project appears not actively maintained. 

Turning to the specific QA annotation platforms, we surveyed two tools: \textit{cdQA-annotator} \cite{mikaelian2019cdqa} and \textit{QA-Turk} \cite{fisch2018qaturk}. The former is a part of a larger web-based QA suite. The annotation workflow in cdQA-annotator allows for direct selection of the answer span, which facilitates the task. Furthermore, it uses the SQuADv1.1 file format for input and output files. Nonetheless, it includes neither contribution nor user management capabilities. Conversely, \textit{QA-Turk}, based on \textit{ieturk} \cite{quach2018ieturk}, is an add-on allowing to create question-answers pairs. By default, it uses  Amazon Mechanical Turk as a crowd-sourcing back-end, but can be used with a local-based alternative. While also tailored for crowd-sourcing, it does not support different roles for the contributors. We note that, while these two platforms do not satisfy our requirements, they include valuable features such as Mechanical Turk integration for \textit{QA-Turk} or the QA tools included in the cdQA suite (model training, visualization, exploration). Finally, those tools do not seem to allow collection of additional answers for a subset of the data (as mentioned above, this allows to both make the evaluation data more robust, and to compute human performance).

PIAFAnno aims to overcome these limitations, specifically in two  areas: dealing with multiple, role-diverse contributors and  making it as easy as possible for them to create annotated samples. As said before, due to the constraints of our protocol, having two types of users (certified and open, see  Section \ref{sec:Protocol}) and keeping volunteering contributors engaged is essential for the success of our approach. These two features can be easily be extended to other annotation efforts, making PIAFAnno a valuable resource for the creation of custom QA datasets. Source code and installation/deployment instructions for PIAFAnno can be found at \url{https://github.com/etalab/piaf}. 

In the following section, we provide a general description of the inner workings of our annotation platform.

\subsection{PIAFAnno System Architecture}

\begin{figure}[h!]
	\centering
	\includegraphics[width=.9\linewidth]{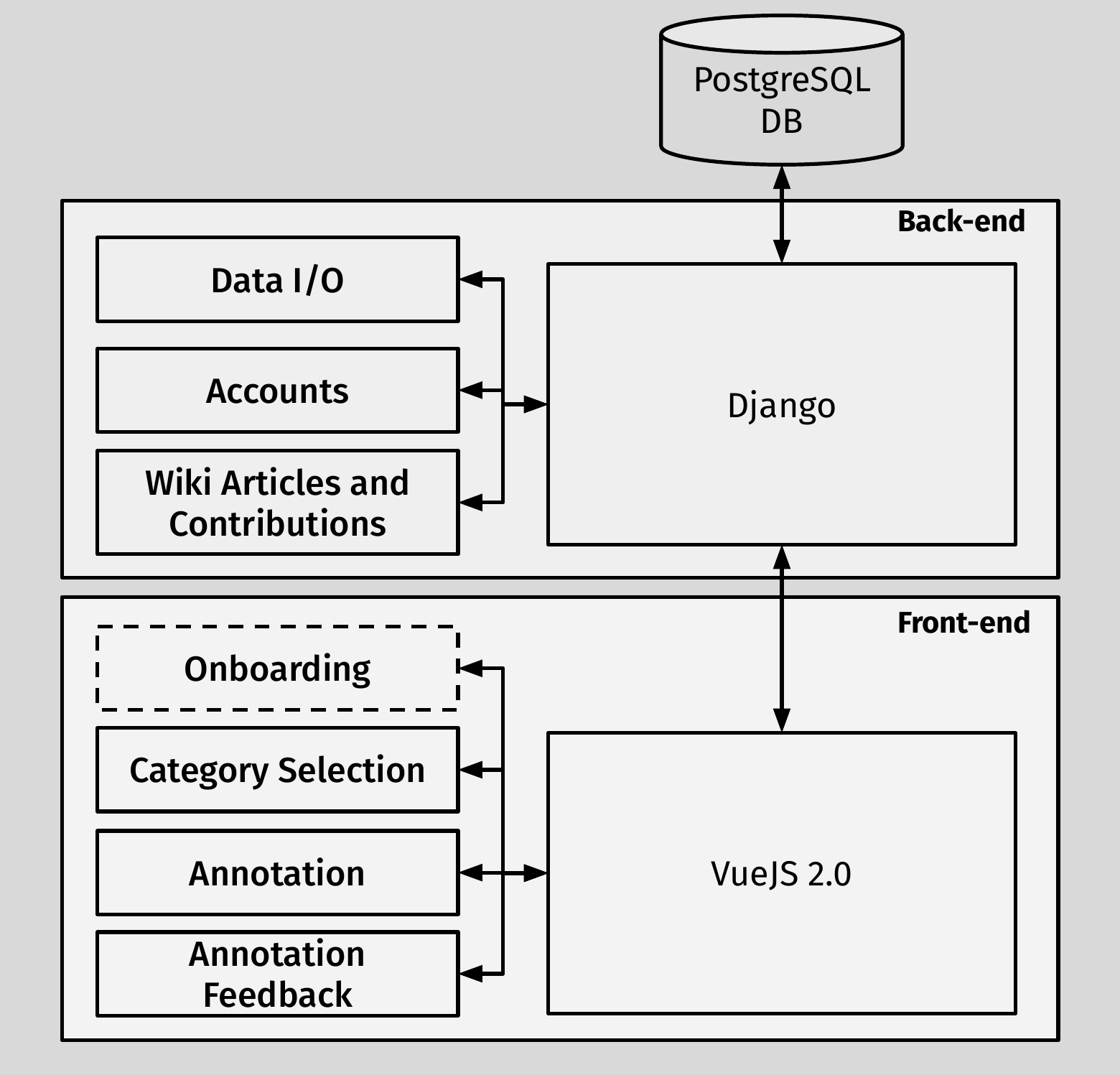}
	\caption{PIAFAnno system architecture.}
	\label{fig:piafarchi}
\end{figure}

The architecture of PIAFAnno is presented in Figure \ref{fig:piafarchi}. The system consists of seven modules attached to either the back-end or the front-end. All the functionalities described in this section are accessible through a web interface. Back-end administration is handled via the Python Django framework.\footnote{\url{https://www.djangoproject.com/}} The front-end annotation modules were developed using the Vuetify\footnote{\url{https://vuetifyjs.com/}} framework on top of VueJS 2.0\footnote{\url{https://vuejs.org/}}.

PIAFAnno is built in a modular fashion, with a focus on simplicity and ease-of-use, and allows for ad-hoc modifications and extensions. Furthermore, the platform is dockerized via \texttt{docker-compose} allowing for quick and straight-forward installation. 

Below we describe the characteristics of each of the seven modules. We cover fist the back-end and then move to the front-end modules.

\paragraph{Data I/O}

The data I/O interface allows for import/export of documents (in our case, French Wikipedia articles and their paragraphs) to be used during the QA collection. Both tasks require and produce JSON files that follow the SQuADv1.1 file format.

\paragraph{Accounts}

This interface allows the platform administrators to manage the contributors' accounts. These include creation date, user type (either regular, admin or super-admin\footnote{The difference between an admin and a super-admin is that the admin is restricted on most of the deletion, modification, and insertion database operations.}), their role (certified or open). Admin users can modify the certified status of the contributors of the platform. Via this module, administrators can access the scores obtained by the contributors, and are thus able to fine-tune the process for future onboardings or to change a contributor status accordingly.

Once an account is created for a contributor, they receive a validation link via email, allowing them to login into the platform.
Standard facilities such as password reset are also managed.

\paragraph{Wikipedia Articles and Contributions}

This module allows to explore the source texts\footnote{In this study, these are Wikipedia articles. Other types of text sources may be used as input.} as well as to monitor the contributions. Details on the created samples (timestamp, author, content, etc.) are available for inspection. This monitoring interface allows for real-time monitoring of the data collection effort.

Similarly, it allows fine-granularity access to the source data, providing article-level details such as category (or topic), and the contributor role required for annotation. At paragraph-level, besides the inherited information, contents and annotation status (e.g. whether question/answer pairs have already been generated on it) are made available.

Below, we move to describe the interfaces the contributors directly interact with, during the annotation.

\paragraph{Onboarding}

The onboarding module aims to facilitate the comprehension of the platform. During preliminary user tests, we found that contributors need a set of guidelines to begin the annotation process according to our quality requirements (e.g., question complexity, answer span length, and so on). It also include a short assessment to evaluate the level of understanding of the annotation guidelines for the onboarding user. The procedure thus consists of two steps: i) the explanation of the guidelines, and ii) the user assessment via simple evaluation questions. This allows us to limit the number of bots, trolls, and  contributors who misunderstand the guidelines of the annotation process. This module is still under active development.

\paragraph{Category Selection}

As described in Section \ref{sec:Protocol}, we manually categorized the source Wikipedia articles into seven topics. This interface allows users to select the category they would like to contribute samples on.

\paragraph{Annotation}

\begin{figure}[t!]
	\centering
	\includegraphics[width=1\linewidth]{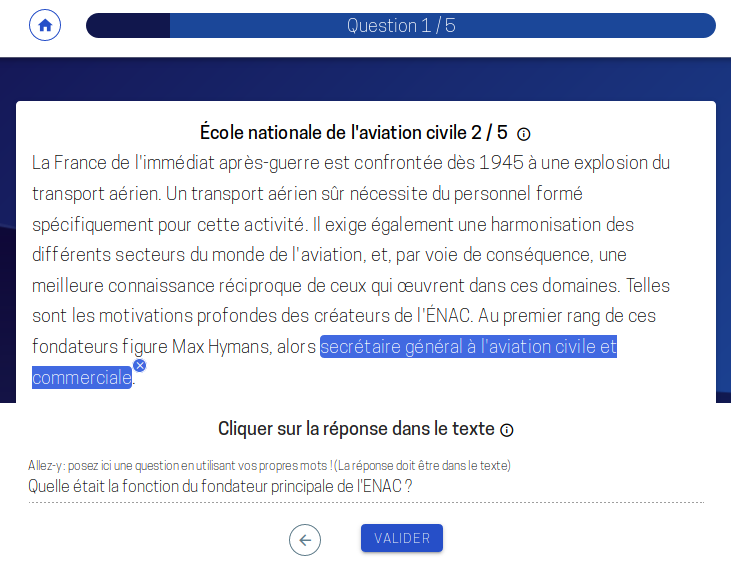}
	\caption{PIAFAnno's annotation page.}
	\label{fig:qainterface}
\end{figure}

Users are required to come up with five question-answer pairs per paragraph. The paragraph is shown according to the original order in their respective Wikipedia article (see Figure \ref{fig:qainterface}). The article shown, and thus its paragraphs, belong to the category previously chosen by the contributor.

The annotation procedure consists of three steps:
\begin{enumerate}
\item \textit{Paragraph reading}: the main area of the page is dedicated to the paragraph from where contributors will create their annotations. The text formatting (e.g. font, spacing, line height, color contrast) has been reviewed by a design specialist. A progress bar allows contributors to know where they are in the current annotation task. 

\item \textit{Question writing}: the question input field is placed below the reading area. The input is limited to 200 characters.

\item \textit{Answer selection}: within the paragraph reading area, the user can select the answer to question they previously wrote. We found that web browsers' traditional text span selection lacks speediness, usability, and regularly shows buggy mobile interaction. Specifically, during our preliminary tests, we found several cases of highlighting of incomplete words, and thus incomplete answers, which is a situation we needed to avoid. To address this problem we developed a custom highlighting component. In this component, the highlighting is done at word-level, rather than at character-level. This ensures complete answers and increases the annotation efficiency. The answer selection works as follows: with the first click, the user  selects the initial word of the answer, while with the second click they will select the last word of the answer span. Finally, the complete answer text span is automatically inferred.
\end{enumerate}

The question-answer pairs can always be edited by the contributor. Once five question-answer pairs are generated for a given paragraph, they are sent to the back-end and stored in the database. The user is then redirected to the annotation feedback page described below.

\paragraph{Annotation Feedback}

The annotation feedback page is shown after a paragraph is completed by the annotator. It congratulates the user and shows them general statistics as a contributor, such as the running number of data samples contributed. This is part of our effort increase and sustain engagement for our contributors.

\paragraph{Additional answers}

The certified users have access to an additional annotation mode. It is similar to the original annotation procedure, but in this case the question is already written and the task is to select the corresponding answer.

This procedure is composed of two steps (as in Annotation mode):
\begin{enumerate}
\item \textit{Paragraph and Question reading},
\item \textit{Answer selection}.
\end{enumerate}

\begin{figure}[t!]
	\centering
	\includegraphics[width=1\linewidth]{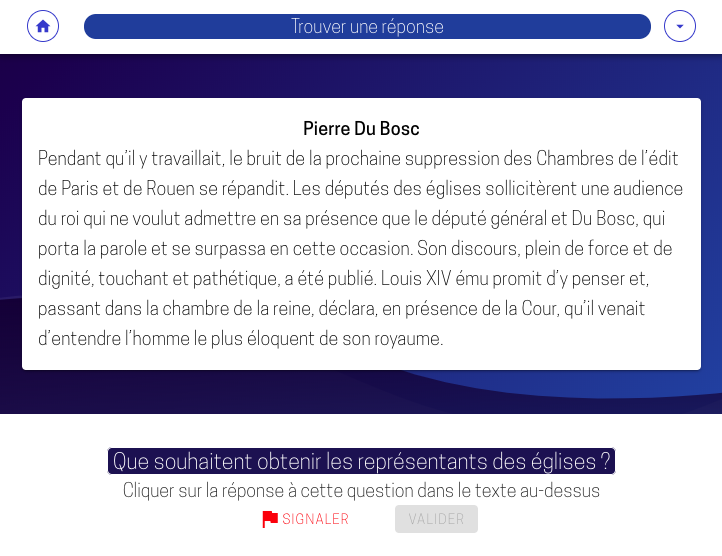}
	\caption{PIAFAnno's additional answer page.}
	\label{fig:qainterface2}
\end{figure}

The question comes from the pool of question-answer pairs already created by other contributors. We consider a sample as complete when two additional answers are selected. In the end, every question will have three answers from three different contributors. Additionally, the user can if needed flag questions (e.g. as unanswerable, ambiguous or offensive questions) using a dedicated button. A screenshot of this mode can be seen in Figure \ref{fig:qainterface2}. Additional answers are, at the time of this writing, under collection, and will be included in the upcoming 1.1 release of the dataset.

\section{Participatory Approach}
\label{sec:crowdsourcing}
We base our approach on the protocol described previously (see Section \ref{sec:Protocol}). We introduce necessary modifications due to the constraints we set to work with: notably, not using Mechanical Turk or similar platforms, thus exploring a method to source voluntary contributions while using a dedicated annotation platform.

The reasons for not using Mechanical Turk (or similar platforms) derive from the participatory approach we adopted. Given its experimental nature, we want to have full control on the entire process and to keep it as fair as possible to the volunteering contributors. Full ownership of the process and the associated software tools allows us to quickly and easily modify the annotation effort methodology ``on-the-fly": the required changes can be implemented and deployed as transparently as possible for the contributors, safeguarding a pleasant annotation experience. 

The creation of quality question-answer pairs is a complex task that, as noticed in our first annotation events, requires a considerable amount of concentration and engagement. Hence, the annotation procedure needs to keep the users motivated and engaged to discover and continue to use the platform. At the same time, we need to maximize the number of contributions by rallying volunteer contributors. As discussed in the previous sections, the developed software includes features aiming at facilitating the annotation task and at engaging participation. On the other hand, to entice the participation of contributors, as a first approach we explore a voluntary crowdsourced acquisition method, inspired by the Mozilla CommonVoice project. 

Specifically, we organize weekly annotation events, or \textit{annotathons}, where contributors are invited to come and participate in the creation of the French QA dataset. These sessions are guided by the PIAF team in order to prepare the annotators to use the annotation tool and to generate high-quality annotations. Since participants receive in-house training into the flow and quality expectations of the annotation process, we consider them as certified annotators (as explained before). The first contributors for these sessions were mainly drawn from the public sector.\footnote{We leverage Etalab's network and expertise on the organization of annotathons. Etalab is the French government Open Data task force.}

The layers of contributions we are currently testing are the following:

\begin{enumerate}
\item \textit{Neighboring}: we promote the weekly annotation event across our neighboring ecosystem network,
\item \textit{Traveling}: we go and present our annotation project to local natural language processing, artificial intelligence or open data related events,
\item \textit{Peripheral}: we build links with and among the external scientific and/or activist communities to get feedback and new avenues of collaboration in order to increase the project's relevance,
\item \textit{International}: as this project is a French-language effort, and not only French (from France), we seek contributions with other governments, research laboratories, institutions, and other communities beyond the French borders.
\end{enumerate}

Our voluntary contribution approach is currently being tested. Other means of contribution may be explored. We also note that during all the annotation events we carry out, we put emphasis on enhancing artificial intelligence literacy through short, educational presentations. Furthermore, since these crowdsourcing events are innovative within the French public-sector context, we are working with sociology specialists in order to observe and provide feedback on the dynamics and development of the project.

\section{Dataset Analysis}
\label{sec:analysis}

\begin{table*}
\begin{center}
    \begin{tabular}{ p{6em} p{13em} p{20em} r}
    \toprule
    Reasoning &Description & Example & Percentage \\
    \hline
    Synonymy & Major correspondences between the question and the  answer sentence are synonyms. & 
    Q: Combien de Polonais \textbf{vivent} sur Terre?
    
    Sentence: Avec une population de \underline{38 millions} d'\textbf{habitants}, la Pologne [...] & 
    41.36 \% 
    \vspace{.5em}\\
    \hline
    World Knowledge & 
    Major correspondences between the question and the answer sentence  require  world  knowledge  to resolve.& 
    Q: Quelle \textbf{maladie} a terrassé le prédécesseur du général de Rochambeau? 
    
    Sen.: Le général Leclerc meurt de \underline{\textbf{la fièvre jaune}}. &
    26.18 \%  
    \vspace{.5em}\\ 
    \hline
    Syntactic variation &
    After the question is paraphrased into declarative form, its  syntactic dependency structure does not match that of the answer  sentence even after local modifications. & 
    Q: Quelle est la profession d'Alain Testart?
    
    Sen.: Mais \textbf{les différents \underline{anthropologues}} qui se qualifient d'évolutionnistes de nos jours, \textbf{tel qu'Alain Testart} et Christophe Darmangeat, proposent [...] & 67.54  \% 
    \vspace{.5em}\\
    \hline
    Multi sentence reasoning & 
    There  is  anaphora, or  higher-level fusion  of  multiple  sentences  is  required.& 
    Q: Les iles greques sont elles toutes habitées?
    
    Sen.: \textbf{La Grèce}, d'une superficie de [...] et partage des frontières maritimes avec [...]. La mer Ionienne à l'ouest et [...], encadrent \textbf{le pays} dont le cinquième du territoire est \textbf{constitué de plus de \underline{9 000 îles et îlots dont près de 200 sont habités}}.
    & 12.04 \% 
    \vspace{.5em}\\
    \hline
    Ambiguous &
    We  don’t  agree  with  the  annotator's  answer,  or  the  question does not have a unique answer.&
    Q: Quel a été le \textbf{résultat} en demi-finale?
    
    Sen.: En demi-finale à Wembley, Arsenal du se débarrasser des tenant du titre, Wigan Athletic. Le match se \textbf{termine sur un \underline{score d'1-1}}. [...] Arsenal se \textbf{qualifia aux tirs au but, 4-2}, avec [...]
    & 9.95 \%\\
    \toprule
    \end{tabular}
\end{center}
\caption{$N=191$  randomly sampled triplets were manually assigned into  one  or  more  of  the  above  categories. Words relevant  to  the  corresponding reasoning type are in bold, and the annotated answer is underlined.}
\label{tab:manual}
\end{table*}

\begin{figure*}
	\centering
	\includegraphics[width=.39\linewidth]{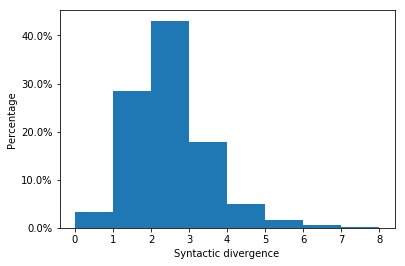}
	\includegraphics[width=.39\linewidth]{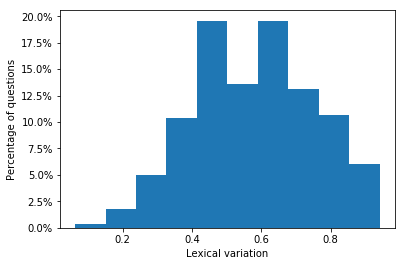}
	\caption{Distributions of syntactic divergence (left) and lexical variation (right) on question/sentence pairs for \emph{PIAFv1.0}.}
	\label{fig:autometrics}
\end{figure*}

\begin{table}
    \centering
    \begin{tabular}{l|c|c}
    $\downarrow train \hphantom{blablabla}eval\rightarrow$& FQUAD & \emph{PIAFv1.0} \\
    \hline
    SQuAD-Fr & 78.39 & 68.90 \\
    \hline
    FQUAD &78.96 & 66.30\\
    \hline
    SQuAD-Fr + FQUAD & 81.09 & 71.11\\
    \hline
    SQuAD-Fr + FQUAD\textsubscript{sub} & 79.12 & 69 \\
    \hline
    SQuAD-Fr + \emph{PIAFv1.0} & 79.48 & -\\
    \hline
    FQUAD + \emph{PIAFv1.0} & 79.64 & -\\
    \hline
    \emph{all} & 82.01 & -\\
    \end{tabular}
    \caption{F1 scores obtained after fine-tuning CamemBERT for QA on different training datasets. Models are evaluated on FQUAD (dev) and \emph{PIAFv1.0} (when applicable). \emph{all} refers to the union (shuffled) of FQUAD (train), SQuAD-Fr (train), and \emph{PIAFv1.0.}}
    \label{tab:results}
\end{table}

\emph{PIAFv1.0} compounds to 3835 question/answer pairs collected on 761 paragraphs under a \emph{certified} setup: volunteers gathered for \emph{annotathons} animated by the PIAF team. 285 volunteers contributed annotated samples to \emph{PIAFv1.0}.

Consistently with SQuAD, we report in Table~\ref{tab:manual}  the human assessment scores for 191 triplets (one per article, randomly sampled). Conversely, Figure~\ref{fig:autometrics} shows the distributions for lexical variation and syntactic divergence among all triplets.

The PIAF project went public on October 3rd, 2019. As mentioned in Section~\ref{sec:related_work}, a similar effort~\cite{d2020fquad} was concurrently being conducted, unbeknownst to us, producing a quantitatively larger French QA dataset named FQUAD. Therefore, we include a comparative analysis of PIAFv1.0 and FQUAD. 

For some analyses below, we use the automatic French translation of the English SQuAD (v1.1) obtained via Google Translate APIs,\footnote{\url{translate.google.com}} referred to as \emph{SQuAD-Fr}. It compounds to 74308 and 9455 samples for training and development sets, respectively, after filtering out bad samples (e.g. those for which the translated answer could not be recovered from the input).

In \cite{d2020fquad}, the authors rely on CamemBERT \cite{camembert} for their evaluations, but do not report the hyper-parameters used. For all our experiments, we use $batch\_size=8$, $learning\_rate=3e^{-5}$, $n\_epochs=2$, $max\_seq\_len=384$, $doc\_stride=128$.

In Table~\ref{tab:results}, we report the results obtained by fine-tuning CamemBERT\footnote{We use the implementation provided at \url{https://github.com/huggingface/transformers}.} on different combinations of the datasets available. First, when evaluated on \emph{PIAFv1.0} data, the performances of all models are significantly lower than on the FQUAD development set. Second, a model trained on the automatic French translation of SQuAD obtains, on the FQUAD development set, the same performance as using the FQUAD training set. 

Turning to using \emph{PIAFv1.0} samples for training, rather than for evaluation, we observe that a model trained on SQuAD-Fr, and augmented with the \emph{PIAFv1.0} samples, obtains comparable results (about half a F1 point improvement) w.r.t. using the FQUAD training set. Note that our CamemBERT finetuned on FQUAD training samples, and evaluated on FQUAD development data, obtains a performance significantly lower (10 absolute F1 points) than that reported by \cite{d2020fquad} on the FQUAD hidden test set.\footnote{The same CamemBERT model, evaluated on the SQuAD-Fr dev set, obtains better results (F1: 73.28, EM: 59.18) than those reported by \cite{d2020fquad} (F1: 70.7, EM: 56.9).} This can be explained by a number of factors, including: hyperparameters' setup (not reported in the FQUAD paper); the use of additional answers for computing evaluation scores on the hidden test set (although, for reference, this factor only justifies 3-4 points of difference on the SQuAD dev set). 

As shown by \cite{geva2019we}, employing a small number of annotators (for FQUAD, $N=18$) can result in annotator bias: if unaccounted for, e.g. by creating training/evaluation splits based also on annotator identifiers, models might fail to generalize to samples produced by annotators that did not contribute to the training set. In \emph{PIAFv1.0} this risk is mitigated by the large number of volunteering contributors ($N=258$). 
Finally, having a larger pool of annotators, we also ensure higher diversity in the textual samples collected: this seems to have a direct impact for training \cite{camembert} and for evaluation robustness.
The results shown in Table~\ref{tab:results} indicate that our participatory approach allows to obtain relatively more challenging evaluation samples. 

Furthermore, taking as reference a model trained on SQuAD-Fr only, we observe that the addition of samples from \emph{PIAFv1.0} during training obtains a slightly larger improvement on FQUAD (dev) than adding a comparable subsample (FQUAD\textsubscript{sub}) from the FQUAD training set. Since in the former case there is no risk of annotator bias (no overlap between FQUAD and PIAF annotators), this shows that \emph{PIAFv1.0} samples can also effectively be used for training data augmentation.

\section{Conclusion}
Motivated by the scarcity of non-English data, we described our ongoing effort towards gathering native QA samples for the French language, using a participatory approach. Rather than a transactional approach to data collection, as usually adopted in crowd-sourcing efforts, we experiment with a comparatively slower and more engaging process, focusing on quality over quantity.
Amongst desirable side-effects of our approach, we highlight the educational aspects, e.g. introducing a wider audience to AI concepts and methodologies during our annothathons. 

The analyses reported indicate that the samples collected in \emph{PIAFv1.0} can effectively be exploited for robust evaluation or training. 

Furthermore, we presented a novel, open-sourced, and language-agnostic data collection platform for Question Answering tasks, developed within the context of Project PIAF. 

Both the PIAFAnno platform and the data collected are released under permissive and open-source licenses, with the goal of engaging a diverse and wide community of practitioners.
We leave the web platform open to the public, in order to collect additional samples from online contributors, which will be included in future releases of the dataset.

\section{Acknowledgements}
The PIAF project is supported by Etalab and receives funding from the Investing for the Future Program, led by the General Secretariat for Investment and the  Deposits and Consignments Fund. 

Our gratitude goes to all participants and previous/current/future contributors to the PIAF campaigns; to Djamé Seddah, Miriam Redi, and to all fellow researchers who provided valuable inputs.

We thank Project Nayuki for granting us permission to adapt and re-use their code, and the Mozilla Common Voice team for their precious feedback regarding the annotation tool and methodology.

\clearpage
\section{Bibliographical References}
\label{main:ref}

\bibliographystyle{lrec}
\bibliography{lrec2020W-xample}


\end{document}